\documentclass{article}

     \PassOptionsToPackage{numbers, compress}{natbib}

\usepackage[final]{neurips_glfrontiers_2022}

\usepackage[utf8]{inputenc} 
\usepackage[T1]{fontenc}    
\usepackage{hyperref}       
\usepackage{url}            
\usepackage{booktabs}       
\usepackage{amsfonts}       
\usepackage{nicefrac}       
\usepackage{microtype}      
\usepackage{xcolor}         
\usepackage{multirow}

\usepackage{amsmath}
\usepackage[pdftex]{graphicx} 
\usepackage{subfig}

\DeclareMathOperator*{\argmax}{argmax}

\title{A Graph Is More Than Its Nodes: Towards \\Structured Uncertainty-Aware Learning on Graphs}

\author{%
	Hans Hao-Hsun Hsu\thanks{Equal contribution} $\ ^1$
	\quad Yuesong Shen\footnotemark[1]  $\ ^{1,2}$
	\quad Daniel Cremers  $^{1,2}$\\
	$^1\ $Technical University of Munich, Germany\\
	$^2\ $Munich Center for Machine Learning, Germany\\
	\texttt{\{hans.hsu, yuesong.shen, cremers\}@tum.de}\\
}

\begin{document}

\maketitle

\begin{abstract}
  Current graph neural networks (GNNs) that tackle node classification on graphs tend to only focus on nodewise scores and are solely evaluated by nodewise metrics. This limits uncertainty estimation on graphs since nodewise marginals do not fully characterize the joint distribution given the graph structure. In this work, we propose novel edgewise metrics, namely the edgewise expected calibration error (ECE) and the agree/disagree ECEs, which provide criteria for uncertainty estimation on graphs beyond the nodewise setting. Our experiments demonstrate that the proposed edgewise metrics can complement the nodewise results and yield additional insights. Moreover, we show that GNN models which consider the structured prediction problem on graphs tend to have better uncertainty estimations, which illustrates the benefit of going beyond the nodewise setting.\footnote{Source code available at \url{https://github.com/hans66hsu/structured_uncertainty_metrics}}
\end{abstract}

\section{Introduction}

Learning from graph-structured data has gained increasing attention from the deep learning community and numerous graph neural networks (GNNs) have been introduced to cope with this new paradigm \citep{kipf2017gcn,velivckovic2018gat,qu2019gmnn,wang2021epfgnn}. Commonly used GNN models like Graph Convolution Network (GCN) \cite{kipf2017gcn} and Graph Attention Network (GAT) \cite{velivckovic2018gat} produce nodewise scores from which good classification results can be achieved.

On the other hand, learning on graphs poses particular challenges for uncertainty estimation. Given the graph structure, the node predictions are no longer independent samples. Instead, we have a structured prediction problem \citep{nowozin2011structured} and should also account for their interdependencies. In other words, high-quality uncertainty estimation for GNNs requires reliable estimation of the joint output distribution. Hence considering only nodewise marginal output distributions is insufficient.

From this structured prediction perspective, we see a considerable limitation of existing works on uncertainty-aware learning for GNNs, namely that they only employ nodewise metrics as evaluation criteria. Given the abundant existing work on uncertainty-aware learning for standard multi-class classification \citep{snoek2019canyoutrust,guo2017calibration,sensoy2018evidentialdl}, many ideas have been adapted to the GNN case. This includes approaches based on Bayesian formulations \citep{zhang2019bayesgcn,hasanzadeh2020graphdropconnect}, evidential learning \citep{zhao2020graphedl,stadler2021gpn}, as well as post-hoc calibration methods \citep{teixeira2019gnncalib,wang2021cagcn,hsu2022what}. However, for estimating the quality of predictive uncertainty, these approaches directly use the nodewise metrics from the multi-class classification literature, which are intended for i.i.d.\ test samples and ignore the graph structure. 

To tackle this issue, we investigate novel uncertainty estimation metrics that consider the interdependencies in the graph structure. Specifically, we propose the edgewise expected calibration error (ECE) which relies on the graph structure and the edgewise marginal distributions. Furthermore, we formulate agree and disagree ECEs to analyze the agreeing and disagreeing edge marginals separately. Our experiments in Section~\ref{sec:exp} show that the proposed edgewise metrics can complement the nodewise ECE results and offer additional insights.
To the best of our knowledge, structured uncertainty estimation metrics have not yet been proposed for graph learning.

Equipped with the proposed edgewise metrics, we compare uncertainty estimation of GNN models which simply produce nodewise scores, e.g., GCN and GAT, with models which explicitly tackle the structured prediction problem, like Graph Markov Neural Network (GMNN) \citep{qu2019gmnn} and Explicit Pairwise Factorized Graph Neural Network (EPFGNN) \citep{wang2021epfgnn}. While naive mean field \citep{nowozin2011structured} can approximate predictions based on nodewise scores to the joint output distribution, we observe in practice that approaches treating graph learning as structured prediction problems tend to have better uncertainty estimation. This demonstrates the benefit of going beyond the nodewise setting, especially for the uncertainty estimation of GNNs.

\begin{figure}[t]
  \centering
  \begin{minipage}[c]{.4\linewidth}
  	\vspace{0pt}
  	\centering
  	\includegraphics[width=\linewidth]{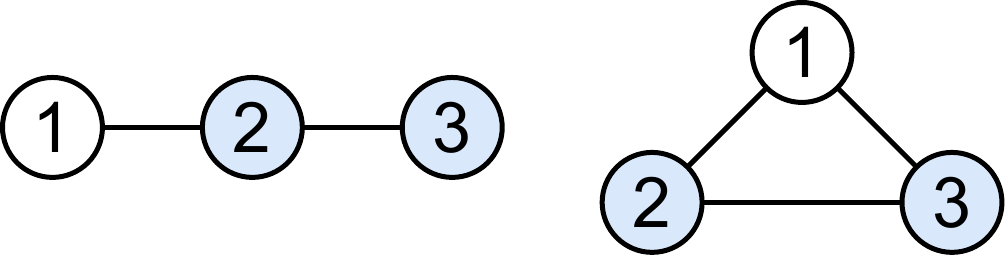}
  \end{minipage}%
  \begin{minipage}[c]{.55\linewidth}
  	\vspace{0pt}
  	\centering
  	\begin{tabular}{cccc}
  		\toprule
  		\multirow{2}{*}{$(p_1; p_2; p_3)$} &\multirow{2}{*}{Nodewise} & \multicolumn{2}{c}{Edgewise} \\
  		 & & Chain & Cycle \\
  		\midrule
  		$(0; 1; 1)$ & 0& 0 & 0 \\
  		$(2/3; 2/3; 2/3)$ & 0& $1/18$ & $1/9$ \\
  		$(0.55; 0.8; 0.7)$ & 0.017& 0 & 0.128 \\
  		\bottomrule
  	\end{tabular}
  \end{minipage}
  \caption{Examples of nodewise and edgewise ECEs for a chain and a cycle. The nodewise ECEs stay the same for both graphs given the same node settings, while the proposed edgewise ECEs can have different values depending on the graph structure. Further details are available in Appendix~\ref{asec:example}.} \label{fig:edge_ece_example}
\end{figure}

\section{Uncertainty estimation on graphs beyond nodewise criteria}

Given a graph $\mathcal{G} = (\mathcal{N}, \mathcal{E})$ with $c$ output classes for each node, the task of node classification aims at predicting the labels $y_U = \{y_i\}_{i \in U}$ of the test nodes $U \subseteq \mathcal{N}$ given the labels  $y_L = \{y_i\}_{i \in L}$ of the training nodes $L = \mathcal{N} \setminus U$ and the input features $x = \{x_i\}_{i \in \mathcal{N}}$. Ideal uncertainty estimation is achieved when the predicted joint distribution $\hat{p}(y_U | x, y_L)$ matches the true joint distribution $p(y_U |x, y_L)$. However, this condition cannot be verified in practice, for the following two reasons:
\begin{enumerate}
	\item It is impossible to estimate $p(y_U |x, y_L)$ with a single set of ground-truth labels $y^*_U$;
	\item The joint distributions are intractable given their exponential size w.r.t.\ the node count.
\end{enumerate}

Instead, one can consider weaker but more tractable alternatives based on marginal distributions. Existing works \citep{teixeira2019gnncalib,wang2021cagcn,hsu2022what} consider the confidence calibration of nodewise marginal distributions and measure the quality of uncertainty estimation using their expected calibration error (ECE) as defined by \citet{guo2017calibration}: Denote $(\hat{y}_i)_{i \in U} = \argmax_{y_U} \hat{p}(y_U | x, y_L)$ the class predictions and $\hat{c}_i = \max_l \hat{p}(y_i=l|x, y_L)$ the confidence of node $i$, we partition the test nodes into $m$ bins $(B^n_1, \dots, B^n_m)$ with $B^n_k = \{j \in U| \frac{k-1}{m} < \hat{c}_j \leq \frac{k}{m} \}$, the nodewise ECE is defined as
\begin{align}
&ECE^{n} = \sum_{k=1}^m \frac{|B^n_k|}{|U|} \big| acc(B^n_k) - conf(B^n_k) \big|, \text{ with } \\
&acc(B^n_k) = \frac{1}{|B^n_k|} \sum_{i\in B^n_k} \mathbf{1}(\hat{y}_i = y^*_i) \text{ and } conf(B^n_k) = \frac{1}{|B^n_k|} \sum_{i\in B^n_k} \hat{c}_i.
\end{align} 

\subsection{Edgewise expected calibration error}

An important drawback of the nodewise ECE is that it neither reflects the dependency w.r.t\ graph structure nor the interdependencies between the nodes. To remedy this, we consider the confidence calibration of edgewise marginals, and propose the following edgewise expected calibration error: Denote $\mathcal{E}_U = \{(i, j) \in \mathcal{E} | i, j \in U \}$ the subset of edges constrained to test nodes, and for edge $(i,j) \in \mathcal{E}_U$ denote $\hat{c}_{i,j} = \max_{l,m} \hat{p}(y_i=l, y_j=m|x, y_L)$ its confidence, we construct a similar edge binning $(B^e_1, \dots, B^e_m)$ with $B^e_k = \{(i,j) \in \mathcal{E}_U| \frac{k-1}{m} < \hat{c}_{i,j} \leq \frac{k}{m} \}$ and define the edgewise ECE as
\begin{align}
&ECE^{e} = \sum_{k=1}^m \frac{|B^e_k|}{|\mathcal{E}_U|} \big| acc(B^e_k) - conf(B^e_k) \big|, \text{ with } \label{eq:edge_ece_1} \\
&acc(B^e_k) = \frac{1}{|B^e_k|} \sum_{(i, j)\in B^e_k} \mathbf{1}\big( \hat{y}_i = y^*_i) \cdot \mathbf{1}\big( \hat{y}_j = y^*_j \big) \text{ and } conf(B^e_k) = \frac{1}{|B^e_k|} \sum_{i\in B^e_k} \hat{c}_{i,j}.  \label{eq:edge_ece_2}
\end{align} 

Figure~\ref{fig:edge_ece_example} provides a concrete example that shows the dependency of edgewise ECE w.r.t\ graph structures. Similar issues also exist for other nodewise metrics like negative log-likelihood (NLL) and Brier score. Their edgewise extensions are also defined in Appendix~\ref{asec:edge_metrics}. 

Along with the quantitative results of edgewise ECE, we can also plot its corresponding reliability diagram \citep{guo2017calibration} for edgewise marginal predictions. They provide additional insights by displaying the over/under-confident tendency of the predictions at each confidence level. 

It is possible to go beyond the edgewise setting and calibrate marginal distributions of larger subgraphs. We focus on the edgewise metrics in this work because it is a minimal extension that accounts for the graph structure and has manageable complexity. Moreover, edge marginals have a special significance for probabilistic graphical models: they determine the exact joint distribution of a tree and correspond to the Bethe approximation in the general case \citep{wainwright2008pgm}.

\subsection{Agree and disagree ECEs}

While the edgewise ECE estimates the general quality of the predicted edge marginals, it can be interesting to consider some additional subsets. Especially, since homophily plays an important role in graph learning \citep{zhu2020h2gcn}, we find it insightful to separate the homophilous and heterophilous cases and consider the subsets of agreeing $\mathcal{E}_A = \{(i, j) \in \mathcal{E}_U | y^*_i=y^*_j \}$ and disagreeing edges $\mathcal{E}_D = \{(i, j) \in \mathcal{E}_U | y^*_i \neq y^*_j \}$. This allows us to define the agree ECE ($ECE^a$) and the disagree ECE ($ECE^d$) similar to Equations~\eqref{eq:edge_ece_1} and \eqref{eq:edge_ece_2} but computed from $\mathcal{E}_A$ and $\mathcal{E}_D$ respectively instead of $\mathcal{E}_U$:
\begin{itemize}
\item To compute the agree ECE, we construct the binning $(B^a_1, \dots, B^a_m)$ for the agreeing edge set $\mathcal{E}_A$ with $B^a_k = \{(i,j) \in \mathcal{E}_A| \frac{k-1}{m} < \hat{c}_{i,j} \leq \frac{k}{m} \}$. The agree ECE is defined as
\begin{align}
&ECE^a = \sum_{k=1}^m \frac{|B^a_k|}{|\mathcal{E}_A|} \big| acc(B^a_k) - conf(B^a_k) \big|, \text{ with } \\
&acc(B^a_k) = \frac{1}{|B^a_k|} \sum_{(i, j)\in B^a_k} \mathbf{1}\big( \hat{y}_i = y^*_i) \cdot \mathbf{1}\big( \hat{y}_j = y^*_j \big) \text{ and } conf(B^a_k) = \frac{1}{|B^a_k|} \sum_{i\in B^a_k} \hat{c}_{i,j}.
\end{align} 
\item Analogously, we construct the binning $(B^d_1, \dots, B^d_m)$ for the disagreeing edge set $\mathcal{E}_D$ with $B^d_k = \{(i,j) \in \mathcal{E}_D| \frac{k-1}{m} < \hat{c}_{i,j} \leq \frac{k}{m} \}$. The disagree ECE is defined as
\begin{align}
&ECE^d = \sum_{k=1}^m \frac{|B^d_k|}{|\mathcal{E}_D|} \big| acc(B^d_k) - conf(B^d_k) \big|, \text{ with } \\
&acc(B^d_k) = \frac{1}{|B^d_k|} \sum_{(i, j)\in B^d_k} \mathbf{1}\big( \hat{y}_i = y^*_i) \cdot \mathbf{1}\big( \hat{y}_j = y^*_j \big) \text{ and } conf(B^d_k) = \frac{1}{|B^d_k|} \sum_{i\in B^d_k} \hat{c}_{i,j}.
\end{align} 
\end{itemize}

\section{Experiments} \label{sec:exp}

We conduct a series of experiments to empirically assess the proposed edgewise metrics and analyze the uncertainty estimation of GNNs. We focus on solving the node classification problem on commonly used benchmark datasets Cora \citep{sen2008coraciteseer}, Citeseer \citep{sen2008coraciteseer}, and Pubmed \citep{namata2012pubmed}. In terms of GNN learning frameworks, we compare commonly used GCN \cite{kipf2017gcn} and GAT \cite{velivckovic2018gat} which directly produce nodewise output scores against GMNN \citep{qu2019gmnn} and EPFGNN \citep{wang2021epfgnn} which explicitly address the structured prediction problem formulation. We analyze the quality of their predictive uncertainty using the nodewise, edgewise, and agree/disagree ECE.

Calculation of the edgewise metrics requires estimating the edgewise marginal distributions. GMNN \citep{qu2019gmnn} and EPFGNN \citep{wang2021epfgnn} model the joint output distribution from which the edge marginals can be inferred. For GCN \cite{kipf2017gcn} and GAT \cite{velivckovic2018gat}, we assume that they represent the joint output distribution as the product of nodewise marginal distributions, i.e., a naive mean field \citep{nowozin2011structured} assumption is made. This means that their predicted edgewise marginals are simply products of the neighboring node marginals. Further details about the experimental setup are provided in Appendix~\ref{asec:detail_exp_setup}.

 We summarize the results in Table~\ref{tab:calib_comparison}. Additional results, including the corresponding test accuracies and plots of reliability diagrams, are included in Appendix~\ref{asec:supp_exp_results}.

\paragraph{Nodewise and edgewise expected calibration errors}

Firstly, we consider the nodewise and edgewise ECE results. Both metrics provide indications for the general quality of uncertainty estimation. From the results in Table~\ref{tab:calib_comparison}, we observe two noticeable findings:

\begin{itemize}
    \item GNNs which address node classification as structured prediction (GMNN and EPFGNN) tend to have better uncertainty estimation, in terms of both nodewise and edgewise ECEs, compared to models like GCN and GAT which simply output nodewise scores.
    \item There is a significant correlation between nodewise and edgewise ECEs. This is not surprising, as good uncertainty estimations should make both metrics small. That said, the edgewise ECE has a different focus w.r.t.\ the nodewise ECE, resulting the performance ranking of the Pubmed at odds with each other.
\end{itemize}

\paragraph{Agree and disagree ECEs}

To further analyze the edgewise probabilistic predictions, we partition the edge set into agreeing and disagreeing subsets. We evaluate the quality of their predictive uncertainty separately using agree and disagree ECEs. 

\begin{itemize}
    \item Structured prediction approaches achieve better agree ECEs and worse disagree ECEs. Given the fact that Cora, Citeseer, and Pubmed are all assortative graphs where neighbors tend to agree with each other, these methods seem to focus on calibrating the agreeing edges to reach good overall uncertainty estimation.
    \item GAT consistently yields the best disagree ECEs. This might be an effect of its attention mechanism which allows for distinct treatments of the agreeing and disagreeing edges. 
    \item Reliability diagrams from Appendix~\ref{asec:supp_exp_results} indicate that predictions on agreeing edges tend to be under-confident. On the contrary, the predictions on disagreeing edges tend to be over-confident given that GNNs fail to produce correct high-confidence predictions.
\end{itemize}

	\begin{table}[t]
		\caption{Uncertainty estimation results using various GNN models on Cora, Citeseer, and Pubmed in terms of nodewise, edgewise, agree and disagree ECEs. For all metrics, lower is better.}
		\label{tab:calib_comparison}
		\vspace{7pt} 
		\centering
		\small
		\begin{tabular}{lcccccc}
			\toprule
			Dataset 
			&Model 
			&Nodewise
			&Edgewise
			&Agree
			&Disagree\\
			\midrule
			\multirow{2}{*}{Cora} 
			&GCN &12.47$\pm$4.37 &16.64$\pm$5.53 &24.18$\pm$5.89 &17.87$\pm$3.23\\
			& GAT &15.27$\pm$3.82 &25.37$\pm$5.34 &33.11$\pm$5.60 &\textbf{13.60$\pm$2.51}\\
			& GMNN &\textbf{4.05$\pm$1.30} &\textbf{4.07$\pm$1.86} &\textbf{9.96$\pm$2.59} &30.78$\pm$3.37\\
			& EPFGNN &7.24$\pm$1.96 &8.66$\pm$3.11 &15.21$\pm$3.55 &25.55$\pm$3.99\\
			\midrule
			\multirow{2}{*}{Citeseer} 
			&GCN &11.34$\pm$8.28 &16.96$\pm$10.46 &34.45$\pm$11.69 &29.80$\pm$7.88\\
			& GAT &16.40$\pm$8.51 &27.63$\pm$9.93 &46.70$\pm$10.88 &\textbf{23.14$\pm$7.43}\\
			& GMNN &\textbf{7.97$\pm$2.04} &\textbf{9.21$\pm$2.62} &\textbf{11.16$\pm$4.22} &54.95$\pm$6.02\\
			& EPFGNN &8.05$\pm$5.33 &11.07$\pm$7.37 &18.91$\pm$12.74 &43.95$\pm$13.16\\
			\midrule
			\multirow{2}{*}{Pubmed} 
			&GCN &7.16$\pm$1.40 &3.34$\pm$1.53 &12.64$\pm$1.68 &38.28$\pm$1.78\\
			& GAT &10.58$\pm$1.98 &18.60$\pm$3.01 &29.66$\pm$3.21 &\textbf{28.18$\pm$2.17}\\
			& GMNN &\textbf{4.24$\pm$0.65} &10.78$\pm$0.77 &\textbf{2.62$\pm$0.53} &62.34$\pm$1.36\\
			& EPFGNN &4.84$\pm$1.33 &\textbf{3.12$\pm$1.41} &10.23$\pm$2.09 &44.75$\pm$2.07\\
			\bottomrule
		\end{tabular}
	\end{table}

\section{Discussions}

In this work, we demonstrate the benefits of going beyond the nodewise setting for better uncertainty-aware graph learning. We discuss the need for structured evaluation metrics and propose edgewise metrics that offer additional insights. And our experiments show that tackling the structured prediction on graphs can help the uncertainty estimation.

Our experiments also highlight the well-known challenge of graph learning in heterophilous scenarios. This is currently an active area of research \citep{zhu2020h2gcn} and existing work is less established. We hope our proposed edgewise metrics can help the related research, and in general bolster the development and evaluation of new approaches for uncertainty-aware learning on graphs.


\begin{ack}
This work was supported by the Munich Center for Machine Learning (MCML) and by the ERC Advanced Grant SIMULACRON. The authors would like to thank the anonymous reviewers for their constructive feedback.
\end{ack}

{\small
\bibliographystyle{abbrvnat}
\bibliography{ref}

\begin{thebibliography}{31}
\providecommand{\natexlab}[1]{#1}
\providecommand{\url}[1]{\texttt{#1}}
\expandafter\ifx\csname urlstyle\endcsname\relax
  \providecommand{\doi}[1]{doi: #1}\else
  \providecommand{\doi}{doi: \begingroup \urlstyle{rm}\Url}\fi

\bibitem[Besag(1975)]{pseudo}
J.~Besag.
\newblock Statistical analysis of non-lattice data.
\newblock \emph{The Statistician}, 24\penalty0 (3):\penalty0 179--195, 1975.

\bibitem[Fey and Lenssen(2019)]{pyg}
M.~Fey and J.~E. Lenssen.
\newblock Fast graph representation learning with {PyTorch Geometric}.
\newblock In \emph{ICLR Workshop on Representation Learning on Graphs and
  Manifolds}, 2019.

\bibitem[Guo et~al.(2017)Guo, Pleiss, Sun, and Weinberger]{guo2017calibration}
C.~Guo, G.~Pleiss, Y.~Sun, and K.~Q. Weinberger.
\newblock On calibration of modern neural networks.
\newblock In \emph{ICML}, 2017.

\bibitem[Hasanzadeh et~al.(2020)Hasanzadeh, Hajiramezanali, Boluki, Zhou,
  Duffield, Narayanan, and Qian]{hasanzadeh2020graphdropconnect}
A.~Hasanzadeh, E.~Hajiramezanali, S.~Boluki, M.~Zhou, N.~Duffield,
  K.~Narayanan, and X.~Qian.
\newblock Bayesian graph neural networks with adaptive connection sampling.
\newblock In \emph{ICML}, 2020.

\bibitem[Hsu et~al.(2022)Hsu, Shen, Tomani, and Cremers]{hsu2022what}
H.~H.-H. Hsu, Y.~Shen, C.~Tomani, and D.~Cremers.
\newblock What makes graph neural networks miscalibrated?
\newblock In \emph{NeurIPS}, 2022.

\bibitem[Kingma and Ba(2014)]{kingma2014method}
D.~P. Kingma and J.~Ba.
\newblock Adam: A method for stochastic optimization.
\newblock In \emph{ICLR}, 2014.

\bibitem[Kipf and Welling(2017)]{kipf2017gcn}
T.~N. Kipf and M.~Welling.
\newblock Semi-supervised classification with graph convolutional networks.
\newblock In \emph{ICLR}, 2017.

\bibitem[Koller and Friedman(2009)]{Koller_PGM}
D.~Koller and N.~Friedman.
\newblock \emph{Probabilistic Graphical Models: Principles and Techniques -
  Adaptive Computation and Machine Learning}.
\newblock The MIT Press, 2009.

\bibitem[Kull et~al.(2019)Kull, Perello~Nieto, K\"{a}ngsepp, Silva~Filho, Song,
  and Flach]{kull2019dirichlet}
M.~Kull, M.~Perello~Nieto, M.~K\"{a}ngsepp, T.~Silva~Filho, H.~Song, and
  P.~Flach.
\newblock Beyond temperature scaling: Obtaining well-calibrated multi-class
  probabilities with dirichlet calibration.
\newblock In \emph{NeurIPS}, 2019.

\bibitem[Kumar et~al.(2019)Kumar, Liang, and Ma]{kumar2019verifiedcalib}
A.~Kumar, P.~S. Liang, and T.~Ma.
\newblock Verified uncertainty calibration.
\newblock In \emph{NeurIPS}, 2019.

\bibitem[Lafferty et~al.(2001)Lafferty, McCallum, and Pereira]{Lafferty2001CRF}
J.~D. Lafferty, A.~McCallum, and F.~C.~N. Pereira.
\newblock Conditional random fields: Probabilistic models for segmenting and
  labeling sequence data.
\newblock In \emph{ICML}, 2001.

\bibitem[Murphy et~al.(1999)Murphy, Weiss, and Jordan]{lbp99murphy}
K.~P. Murphy, Y.~Weiss, and M.~I. Jordan.
\newblock Loopy belief propagation for approximate inference: An empirical
  study.
\newblock In \emph{UAI}, 1999.

\bibitem[Namata et~al.(2012)Namata, London, Getoor, and
  Huang]{namata2012pubmed}
G.~Namata, B.~London, L.~Getoor, and B.~Huang.
\newblock Query-driven active surveying for collective classification.
\newblock In \emph{ICML Workshop}, 2012.

\bibitem[Neal and Hinton(1998)]{Neal1998}
R.~M. Neal and G.~E. Hinton.
\newblock \emph{A View of the Em Algorithm that Justifies Incremental, Sparse,
  and other Variants}, pages 355--368.
\newblock Springer, 1998.

\bibitem[Nowozin and Lampert(2011)]{nowozin2011structured}
S.~Nowozin and C.~H. Lampert.
\newblock Structured learning and prediction in computer vision.
\newblock \emph{Foundations and Trends{\textregistered} in Computer Graphics
  and Vision}, 6\penalty0 (3--4):\penalty0 185--365, 2011.

\bibitem[Paszke et~al.(2019)Paszke, Gross, Massa, Lerer, Bradbury, Chanan,
  Killeen, Lin, Gimelshein, Antiga, Desmaison, Kopf, Yang, DeVito, Raison,
  Tejani, Chilamkurthy, Steiner, Fang, Bai, and Chintala]{pytorch}
A.~Paszke, S.~Gross, F.~Massa, A.~Lerer, J.~Bradbury, G.~Chanan, T.~Killeen,
  Z.~Lin, N.~Gimelshein, L.~Antiga, A.~Desmaison, A.~Kopf, E.~Yang, Z.~DeVito,
  M.~Raison, A.~Tejani, S.~Chilamkurthy, B.~Steiner, L.~Fang, J.~Bai, and
  S.~Chintala.
\newblock Pytorch: An imperative style, high-performance deep learning library.
\newblock In \emph{NeurIPS}, 2019.

\bibitem[Qu et~al.(2019)Qu, Bengio, and Tang]{qu2019gmnn}
M.~Qu, Y.~Bengio, and J.~Tang.
\newblock {GMNN}: Graph {M}arkov neural networks.
\newblock In \emph{ICML}, 2019.

\bibitem[Sen et~al.(2008)Sen, Namata, Bilgic, Getoor, Gallagher, and
  Eliassi{-}Rad]{sen2008coraciteseer}
P.~Sen, G.~Namata, M.~Bilgic, L.~Getoor, B.~Gallagher, and T.~Eliassi{-}Rad.
\newblock Collective classification in network data.
\newblock \emph{{AI} Mag.}, 29\penalty0 (3):\penalty0 93--106, 2008.

\bibitem[Sensoy et~al.(2018)Sensoy, Kaplan, and
  Kandemir]{sensoy2018evidentialdl}
M.~Sensoy, L.~M. Kaplan, and M.~Kandemir.
\newblock Evidential deep learning to quantify classification uncertainty.
\newblock In \emph{NeurIPS}, 2018.

\bibitem[Snoek et~al.(2019)Snoek, Ovadia, Fertig, Lakshminarayanan, Nowozin,
  Sculley, Dillon, Ren, and Nado]{snoek2019canyoutrust}
J.~Snoek, Y.~Ovadia, E.~Fertig, B.~Lakshminarayanan, S.~Nowozin, D.~Sculley,
  J.~V. Dillon, J.~Ren, and Z.~Nado.
\newblock Can you trust your model's uncertainty? evaluating predictive
  uncertainty under dataset shift.
\newblock In \emph{NeurIPS}, 2019.

\bibitem[Srivastava et~al.(2014)Srivastava, Hinton, Krizhevsky, Sutskever, and
  Salakhutdinov]{JMLR:v15:srivastava14a}
N.~Srivastava, G.~Hinton, A.~Krizhevsky, I.~Sutskever, and R.~Salakhutdinov.
\newblock Dropout: A simple way to prevent neural networks from overfitting.
\newblock \emph{Journal of Machine Learning Research}, 15\penalty0
  (56):\penalty0 1929--1958, 2014.

\bibitem[Stadler et~al.(2021)Stadler, Charpentier, Geisler, Z{\"u}gner, and
  G{\"u}nnemann]{stadler2021gpn}
M.~Stadler, B.~Charpentier, S.~Geisler, D.~Z{\"u}gner, and S.~G{\"u}nnemann.
\newblock Graph posterior network: Bayesian predictive uncertainty for node
  classification.
\newblock In \emph{NeurIPS}, 2021.

\bibitem[Sutton and McCallum(2009)]{piecewise}
C.~Sutton and A.~McCallum.
\newblock Piecewise training for structured prediction.
\newblock \emph{Machine Learning}, 77\penalty0 (2-3):\penalty0 165, 2009.

\bibitem[Teixeira et~al.(2019)Teixeira, Jalaian, and
  Ribeiro]{teixeira2019gnncalib}
L.~Teixeira, B.~Jalaian, and B.~Ribeiro.
\newblock Are graph neural networks miscalibrated?
\newblock In \emph{ICML Workshop}, 2019.

\bibitem[Veli{\v{c}}kovi{\'c} et~al.(2018)Veli{\v{c}}kovi{\'c}, Cucurull,
  Casanova, Romero, Lio, and Bengio]{velivckovic2018gat}
P.~Veli{\v{c}}kovi{\'c}, G.~Cucurull, A.~Casanova, A.~Romero, P.~Lio, and
  Y.~Bengio.
\newblock Graph attention networks.
\newblock In \emph{ICLR}, 2018.

\bibitem[Wainwright and Jordan(2008)]{wainwright2008pgm}
M.~J. Wainwright and M.~I. Jordan.
\newblock Graphical models, exponential families, and variational inference.
\newblock \emph{Foundations and Trends in Machine Learning}, 1\penalty0
  (1–2):\penalty0 1--305, 2008.

\bibitem[Wang et~al.(2021{\natexlab{a}})Wang, Liu, Shi, and
  Yang]{wang2021cagcn}
X.~Wang, H.~Liu, C.~Shi, and C.~Yang.
\newblock Be confident! towards trustworthy graph neural networks via
  confidence calibration.
\newblock In \emph{NeurIPS}, 2021{\natexlab{a}}.

\bibitem[Wang et~al.(2021{\natexlab{b}})Wang, Shen, and
  Cremers]{wang2021epfgnn}
Y.~Wang, Y.~Shen, and D.~Cremers.
\newblock Explicit pairwise factorized graph neural network for semi-supervised
  node classification.
\newblock In \emph{UAI}, 2021{\natexlab{b}}.

\bibitem[Zhang et~al.(2019)Zhang, Pal, Coates, and
  {\"{U}}stebay]{zhang2019bayesgcn}
Y.~Zhang, S.~Pal, M.~Coates, and D.~{\"{U}}stebay.
\newblock Bayesian graph convolutional neural networks for semi-supervised
  classification.
\newblock In \emph{AAAI}, 2019.

\bibitem[Zhao et~al.(2020)Zhao, Chen, Hu, and Cho]{zhao2020graphedl}
X.~Zhao, F.~Chen, S.~Hu, and J.~Cho.
\newblock Uncertainty aware semi-supervised learning on graph data.
\newblock In \emph{NeurIPS}, 2020.

\bibitem[Zhu et~al.(2020)Zhu, Yan, Zhao, Heimann, Akoglu, and
  Koutra]{zhu2020h2gcn}
J.~Zhu, Y.~Yan, L.~Zhao, M.~Heimann, L.~Akoglu, and D.~Koutra.
\newblock Beyond homophily in graph neural networks: Current limitations and
  effective designs.
\newblock In \emph{NeurIPS}, 2020.

\end{thebibliography}
}

\newpage 

\appendix

\section{Nodewise and edgewise ECEs for concrete examples} \label{asec:example}

To compare nodewise and edgewise ECE and show the dependency of edgewise ECE w.r.t.\ the graph structure, we do a case study for two simple graphs shown in Figure~\ref{fig:edge_ece_example}~Left: a chain and a cycle, both having three binary nodes ($\mathcal{N}=U=\{1,2,3\}$) with labels represented by their white or blue color ($\{w, b\}$), and both having the same ground-truth labels $(y^*_1, y^*_2, y^*_3) = (w, b, b)$. We use the naive mean field approximation and factorize the predicted joint distribution $\hat{p}(y_1, y_2, y_3) = \hat{p}(y_1) \hat{p}(y_2) \hat{p}(y_3)$ into product of nodewise predictions, and denote $p_i = \hat{p}(y_i=b)$ the predicted probability of node $i$ having blue label. In the following, we compute the nodewise and edgewise ECEs using a single bin on the two graphs for the predictions listed in Figure~\ref{fig:edge_ece_example}~Right. As both graphs have the same node settings, their nodewise ECEs are identical for any given prediction.
\begin{description}
\item[$(p_1, p_2, p_3) = (0, 1, 1):$] This is the perfect forecast where the true labels are correctly predicted with full certainty, and we reach zero nodewise and edgewise ECEs for both graphs;  
\item[$(p_1, p_2, p_3) = (\frac{2}{3}, \frac{2}{3}, \frac{2}{3}):$] This is a uniform prediction case that is nodewise calibrated. For both graphs, the label predictions are $(\hat{y}_1,\hat{y}_2,\hat{y}_3) = (b, b, b)$ and the nodewise accuracy and confidence terms are computed as
\begin{equation}
acc(|U|) = \frac{1}{|U|} \sum_{i=1}^3 \mathbf{1}(y^*_i=\hat{y}_i) = \frac{2}{3}; \quad conf(|U|) = \frac{1}{|U|} \sum_{i=1}^3 \hat{p}(\hat{y}_i) = \frac{1}{3} \sum_{i=1}^3 p_i = \frac{2}{3}.
\end{equation}
Thus the nodewise ECE for both graphs is
\begin{equation}
ECE^n = \frac{|U|}{|U|} \big| acc(U) - conf(U) \big| = 0. 
\end{equation}
For the chain case, the edgewise predictions are wrong for edge $(1,2)$ and right for $(2, 3)$, and both edgewise predictions have confidence $\frac{2}{3} \times \frac{2}{3} = \frac{4}{9}$, thus its edgewise ECE is
\begin{equation}
ECE^e(Chain) = \frac{|\mathcal{E}_U|}{|\mathcal{E}_U|} \big| acc(\mathcal{E}_U) - conf(\mathcal{E}_U) \big| = |\frac{1}{2} - \frac{4}{9}| = \frac{1}{18}.
\end{equation}
For the cycle case, all three edges $(1,2), (1,3), (2,3)$ have confidence $\frac{4}{9}$ and only the edge $(2,3)$ has correct label prediction, thus its edgewise ECE is
\begin{equation}
ECE^e(Cycle) = \frac{|\mathcal{E}_U|}{|\mathcal{E}_U|} \big| acc(\mathcal{E}_U) - conf(\mathcal{E}_U) \big| = |\frac{1}{3} - \frac{4}{9}| = \frac{1}{9}.
\end{equation}
\item[$(p_1, p_2, p_3) = (0.55, 0.8, 0.7):$] This is a non-uniform prediction that has zero edgewise ECE on the chain and nonzero edgewise ECE on the cycle.  For both graphs, the label predictions are again $(\hat{y}_1,\hat{y}_2,\hat{y}_3) = (b, b, b)$ and their common nodewise ECE is computed as
\begin{equation}
ECE^n = \Big| \frac{2}{3} - \frac{0.55+0.8+0.7}{3} \Big| \approx 0.017. 
\end{equation}
For the chain case, one out of the two edges is correctly predicted, and its edgewise ECE is
\begin{equation}
ECE^e(Chain)  = \Big| \frac{1}{2} - \frac{0.55 \times 0.8 + 0.8 \times 0.7}{2} \Big| = 0,
\end{equation}
while for the cycle case, one out of its three edges is correctly predicted and its edgewise ECE is computed as
\begin{equation}
ECE^e(Cycle) = \Big| \frac{1}{3} - \frac{0.55 \times 0.8 + 0.8 \times 0.7 + 0.55 \times 0.7}{3} \Big| \approx 0.128.
\end{equation}
\end{description}
The above examples highlight the difference between nodewise and edgewise ECE. Particularly, the edgewise ECEs depend on the graph structure while nodewise ECEs are structure agnostic and only depend on the node settings.



\section{Definitions of other edgewise metrics} \label{asec:edge_metrics} 

Apart from ECE, other edgewise metrics can also be formulated. We provide the formulations for some common metrics in this section.
\begin{description}
\item[Accuracy] While the commonly used nodewise accuracy is defined as
\begin{equation}
ACC^n = \frac{1}{|U|} \sum_{i \in U} \mathbf{1}(\hat{y}_i = y^*_i),
\end{equation}
we can also define the edgewise accuracy 
\begin{equation}
ACC^e = \frac{1}{|\mathcal{E}_U|} \sum_{(i,j) \in\mathcal{E}_U} \mathbf{1}(\hat{y}_i = y^*_i) \cdot \mathbf{1}(\hat{y}_j = y^*_j)
\end{equation}
which computes the ratio of edges with correct label predictions.
\item[Negative log-likelihood] The nodewise negative log-likelihood (NLL), which recovers the joint NLL divided by $|U|$ under the naive mean field assumption, is defined as
\begin{equation}
NLL^n = - \frac{1}{|U|} \sum_{i \in U} \log \hat{p}(y^*_i | x, y_L).
\end{equation}
Similarly, we can define the edgewise NLL as
\begin{equation}
NLL^e = - \frac{1}{|\mathcal{E}_U|} \sum_{(i,j) \in\mathcal{E}_U} \log \hat{p}(y^*_i, y^*_j | x, y_L).
\end{equation}
\item[Brier score] 
The nodewise Brier score has the definition
\begin{equation}
Brier^n = \frac{1}{|U|} \sum_{i \in U} \Big( \big( 1 - \hat{p}(y^*_i | x, y_L) \big)^2 + \sum_{y_i \neq y^*_i} \hat{p}(y_i | x, y_L)^2 \Big),
\end{equation}
and we can formulate the edgewise Brier score as
\begin{equation}
Brier^e = \frac{1}{|\mathcal{E}_U|} \sum_{(i,j) \in\mathcal{E}_U} \Big( \big( 1 - \hat{p}(y^*_i, y^*_j | x, y_L) \big)^2 + \sum_{(y_i, y_j) \neq (y^*_i, y^*_j)} \hat{p}(y_i, y_j | x, y_L)^2 \Big).
\end{equation}
\end{description}

Note that the agree/disagree variants of the above edgewise metrics can be derived analogously by replacing $\mathcal{E}_U$ with $\mathcal{E}_A$ and $\mathcal{E}_D$ respectively. Moreover, while we focus on the ECE metric from \citet{guo2017calibration}, other variants of ECE exist \citep{kumar2019verifiedcalib} and their edgewise and agree/disagree variants can be formulated in a similar manner.

\section{Details of experimental setup} \label{asec:detail_exp_setup}

\subsection{Dataset statistics} \label{subsec:dataset_stat}

	\begin{table}[ht!]
		\caption{Dataset statistics. $K$ index indicates the percentage of test nodes kept in evaluated edges (see Eq. \ref{eq:k_index}). Note that the edge homophily, $K(\mathcal{E}_U)$, $K(\mathcal{E}_A)$ and $K(\mathcal{E}_D)$ depend on how the test nodes are randomly selected and may vary in different splittings. We report the mean and the standard deviation (in percentage) of these values from the splits used in our experiments.}
		\label{tab:data_stat}
		\vspace{7pt} 
		\centering
		\small
		\begin{tabular}{lcccccccc}
			\toprule
			Dataset 
			&Nodes 
			&Edges
			&Features
			&Classes
			&Homophily
			&$K(\mathcal{E}_U)$
			&$K(\mathcal{E}_A)$
			&$K(\mathcal{E}_D)$\\
			\midrule
			Cora &2,708 &10,556 &1,433 &7 
			&80.63$\pm$0.81
			&92.42$\pm$0.37 &84.96$\pm$0.67 &28.77$\pm$0.99\\
			Citeseer &3,327 &9,104 &3,703 &6 &72.63$\pm$0.83
			&85.91$\pm$0.78
			&68.21$\pm$0.82
			&33.76$\pm$0.33\\
			Pubmed &19,717 &88,648 &500 &3 &80.38$\pm$0.28
			&85.97$\pm$0.85
			&74.80$\pm$0.74
			&29.12$\pm$0.52\\
			\bottomrule
		\end{tabular}
	\end{table}
	
We evaluate our models on common citation networks: Cora \citep{sen2008coraciteseer}, Citeseer \citep{sen2008coraciteseer}, and Pubmed \citep{namata2012pubmed}. 
A summary of the dataset statistics is shown in Table \ref{tab:data_stat}.
The edge homophily ratio from \citet{zhu2020h2gcn}
defines the fraction of edges in a graph which connects nodes with the same class label $y^*$.
As we measure the homophily level of the test nodes $U$, the edge homophily ratio is equivalent to the agreeing edges $\mathcal{E}_A$ over test edges $\mathcal{E}_U$.
Its complement is the disagreeing edges that we want to predict.

\begin{equation}
   Homophily = \frac{|\{(i,j):(i,j) \in \mathcal{E}_U\: |\: y^*_i=y^*_j \}|}{|\mathcal{E}_U|} = \frac{|\mathcal{E}_A|}{|\mathcal{E}_U|}.
\end{equation}

We define $\mathcal{E}_U$ by constraining the connected nodes that need to be both test nodes.
If a test node does not have a neighbor which belongs to the test nodes, it does not contribute to the edgewise metrics.
We design an index to indicate the percentage of test nodes kept in $\mathcal{E}_U$:

\begin{equation}
   K(\mathcal{E}_U) = \frac{\text{Number of test nodes which are kept in }\mathcal{E}_U}{|U|}.
   \label{eq:k_index}
\end{equation}

$K(\mathcal{E}_U)$ close to $1$ implies fewer nodes are omitted.
Similarly the index can be applied to $\mathcal{E}_A$ and $\mathcal{E}_D$.

\subsection{Details for model training} \label{subsec:detail_training}

We randomly split the nodes to 15\% as observed and 85\% to unobserved set.
We follow a training framework as \citep{kull2019dirichlet,hsu2022what} to divide the observed set into three-fold where the two portions ($10\%$) are the training set and the other portion ($5\%$) is the validation set.
For all the datasets, we do 5 random data splits, three-fold cross-validation per split, and together with 5 random model initializations.
This results in 75 runs for each experiment.
The models and training framework are implemented in PyTorch \citep{pytorch} and PyTorch Geometric \citep{pyg}.

We choose the commonly used GCN \citep{kipf2017gcn} and GAT \citep{velivckovic2018gat} as the baseline models which do not consider the structured dependency between output nodes. 
For GCN, we use 2 layers with 64 hidden units.
For GAT, we use 2 layers with 8 attention heads, and each consists of 8 hidden units.
Dropout \citep{JMLR:v15:srivastava14a} is applied between layers with dropout rate $0.5$.
We train both models by minimizing NLL using Adam optimizer \citep{kingma2014method}.
The learning rate is set to $0.01$ with the weight decay 5e-4. 
We set the maximum epochs as $2000$ and stop the training if the model does not improve on the validation set over the past $100$ epochs.

For the structured prediction models we choose GMNN \citep{qu2019gmnn} and EPFGNN \citep{wang2021epfgnn}.
Both models are trained with the EM framework \citep{Neal1998} to tackle the partially observed scenario of node classification on graphs and maximize the evidence lower bound instead of the intractable observed log-likelihood.

GMNN consists of one GCN as GNN backbone and another GCN to approximate the Conditional Random Field (CRF) \citep{Lafferty2001CRF}.
Both GCNs have the same hyperparameters as the baseline GCN.
We follow the training pipeline as the original paper. 
We first pre-train the GNN backbone for $200$ epochs
For M step we maximize the pseudolikelihood function \citep{pseudo} and update the approximated CRF with $100$ epochs.
For E step we update the GNN backbone for $100$ epochs to approximate the true joint distribution. 
We observe the approximated CRF can achieve better performance than the GNN backbone by drawing multiple samples.
In practice we draw $32$ samples and compute the mean as the prediction for evaluation.

EPFGNN consists of one GCN backbone and a Markov Random Field (MRF) \citep{Koller_PGM} which models the label dependency explicitly using a shared compatibility matrix.
The GCN backbone produces as output the unary energies of the MRF.
Same as GMNN, we first pretrain the GCN backbone for $200$ epochs.
For M step, we train the GCN backbone and MRF jointly by maximizing piecewise training objective \citep{piecewise}.
Following \citet{wang2021epfgnn}, the training stops if the model does not improve its accuracy on the validation set over $10$ epochs.
For E step, we update a proposal distribution using mean field to approximate the true joint distribution.
To get edge marginals from the MRF, we perform loopy belief propagation \citep{lbp99murphy} with $100$ iterations for evaluation.

\section{Additional experimental results} \label{asec:supp_exp_results}

\begin{figure}[ht]
		\centering
		\setlength{\tabcolsep}{0.02\textwidth} 
		\begin{tabular}{cccc}
			\includegraphics[width=0.20\textwidth]{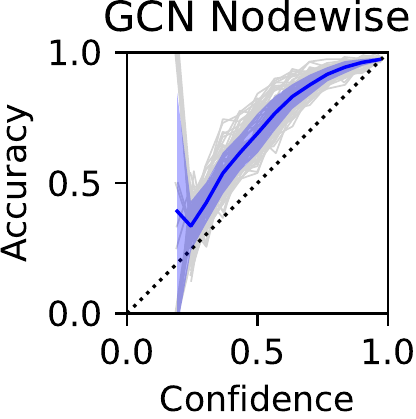} &
			\includegraphics[width=0.20\textwidth]{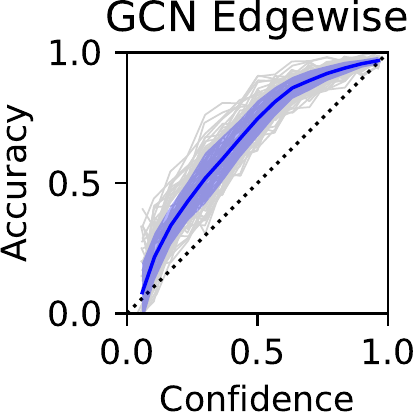} &
			\includegraphics[width=0.20\textwidth]{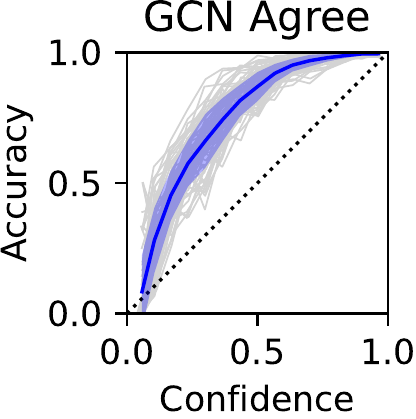} &
			\includegraphics[width=0.20\textwidth]{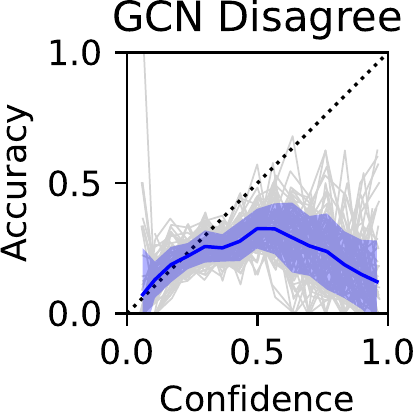} \\
			\includegraphics[width=0.20\textwidth]{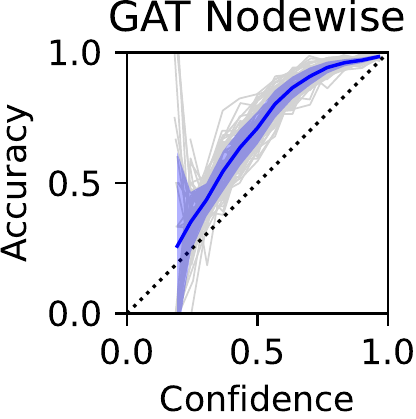} &
			\includegraphics[width=0.20\textwidth]{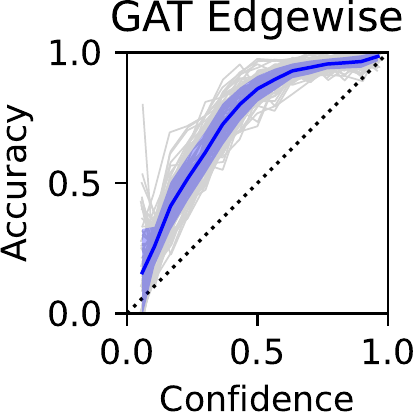} &
			\includegraphics[width=0.20\textwidth]{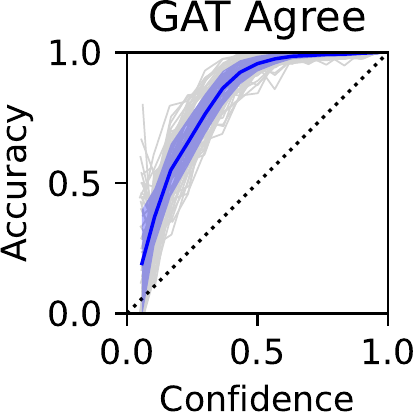} &
			\includegraphics[width=0.20\textwidth]{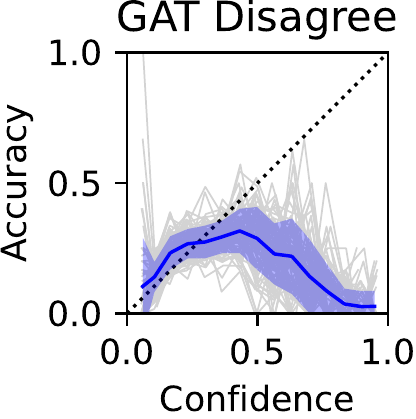} \\
			\includegraphics[width=0.20\textwidth]{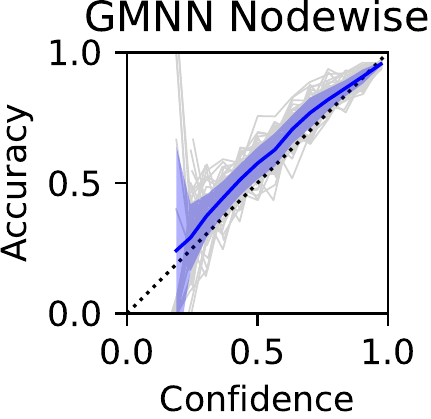} &
			\includegraphics[width=0.20\textwidth]{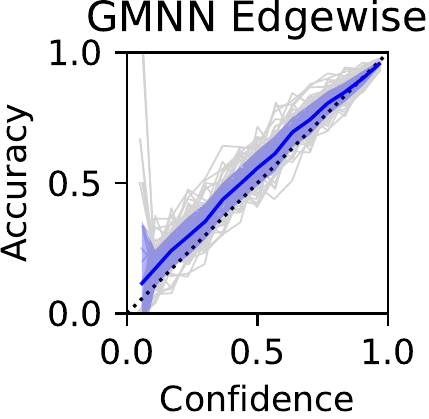} &
			\includegraphics[width=0.20\textwidth]{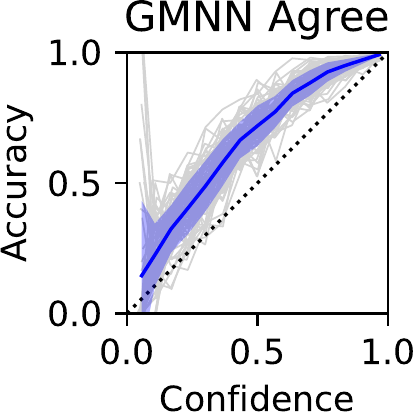} &
			\includegraphics[width=0.20\textwidth]{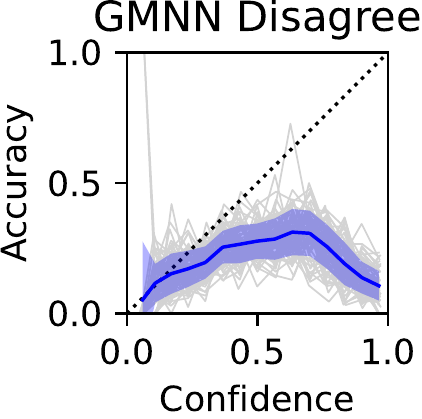} \\
			\includegraphics[width=0.20\textwidth]{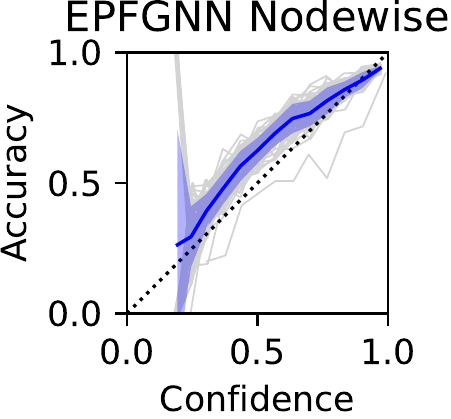} &
			\includegraphics[width=0.20\textwidth]{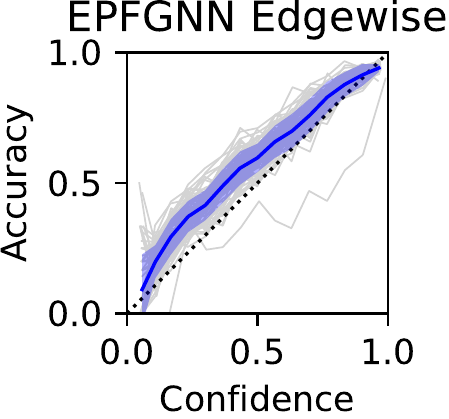} &
			\includegraphics[width=0.20\textwidth]{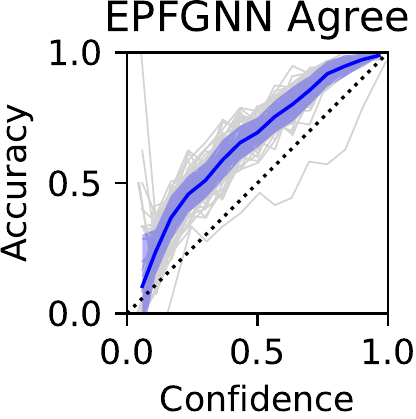} &
			\includegraphics[width=0.20\textwidth]{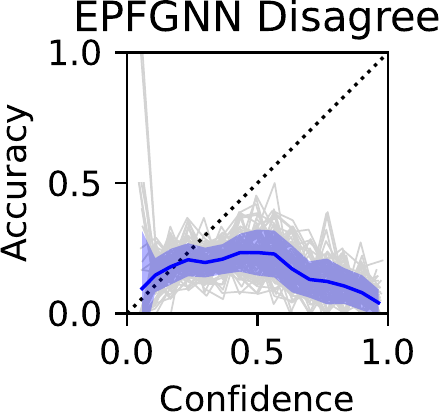} \\
		\end{tabular}
		\caption{Reliability diagrams of GCN, GAT, GMNN, and EPFGNN trained on Cora. The light gray lines are the results of individual experiments. The solid blue line and the shaded blue area are the mean and the standard deviation over 75 runs.} \label{fig:miscalib_tendency_cora}
	\end{figure}

\begin{figure}[ht]
		\centering
		\setlength{\tabcolsep}{0.02\textwidth} 
		\begin{tabular}{cccc}
			\includegraphics[width=0.20\textwidth]{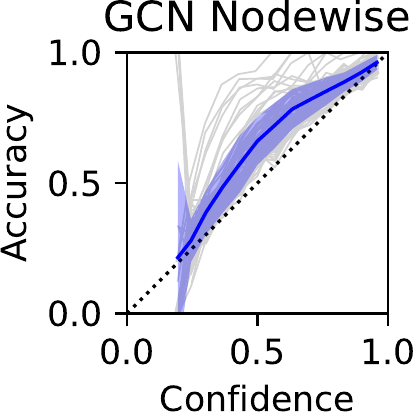} &
			\includegraphics[width=0.20\textwidth]{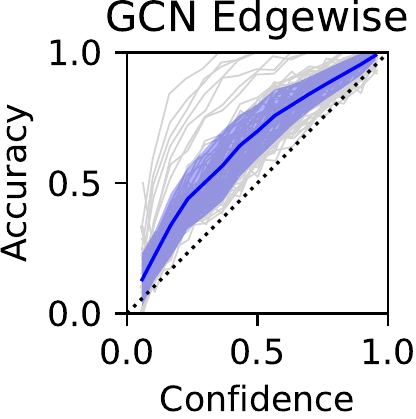} &
			\includegraphics[width=0.20\textwidth]{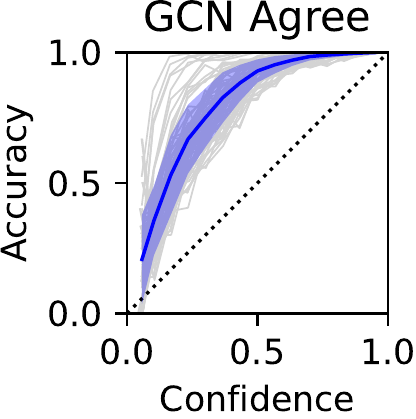} &
			\includegraphics[width=0.20\textwidth]{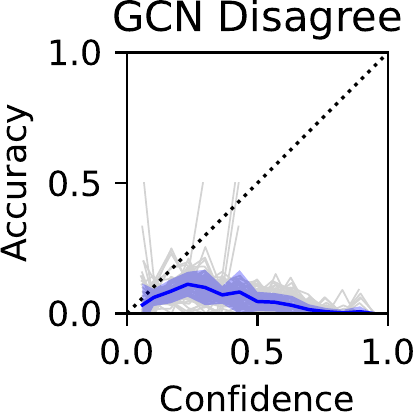} \\
			\includegraphics[width=0.20\textwidth]{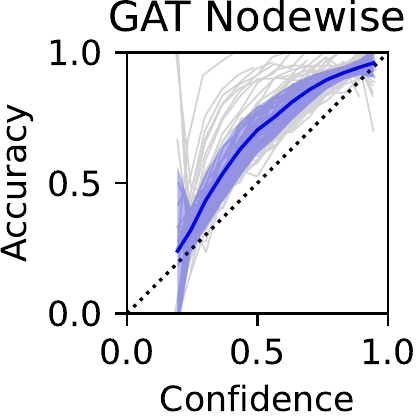} &
			\includegraphics[width=0.20\textwidth]{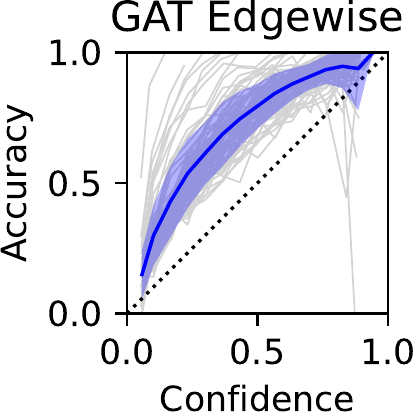} &
			\includegraphics[width=0.20\textwidth]{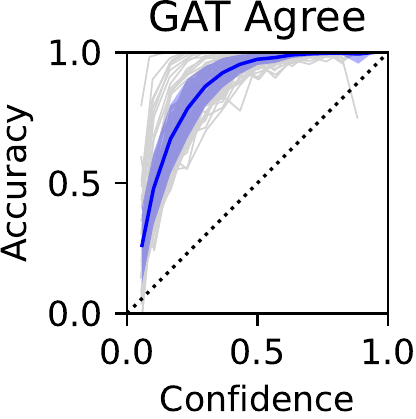} &
			\includegraphics[width=0.20\textwidth]{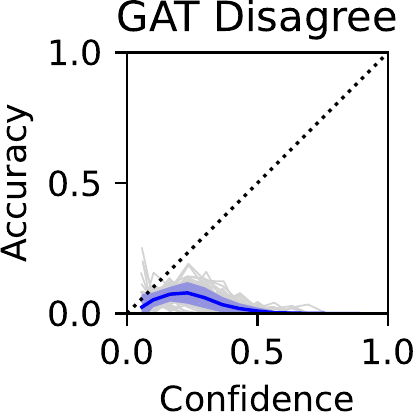} \\
			\includegraphics[width=0.20\textwidth]{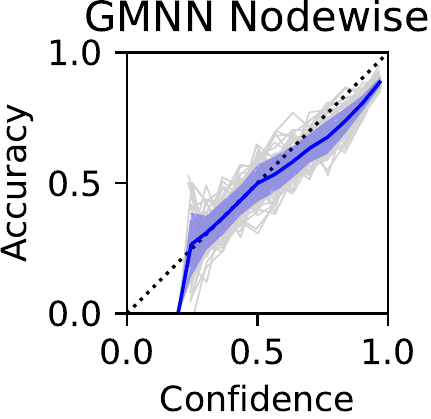} &
			\includegraphics[width=0.20\textwidth]{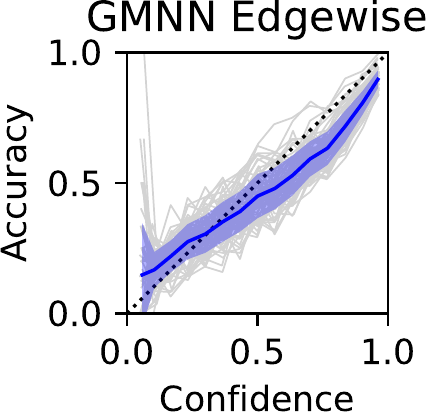} &
			\includegraphics[width=0.20\textwidth]{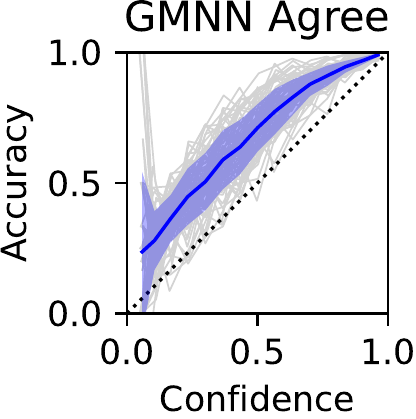} &
			\includegraphics[width=0.20\textwidth]{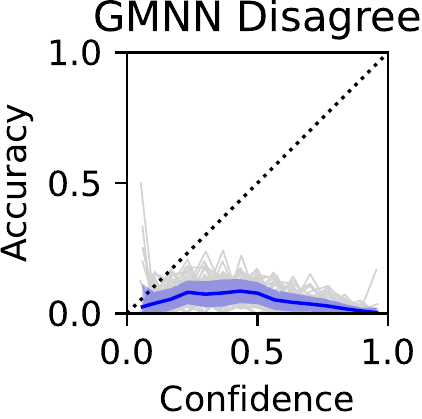} \\
			\includegraphics[width=0.20\textwidth]{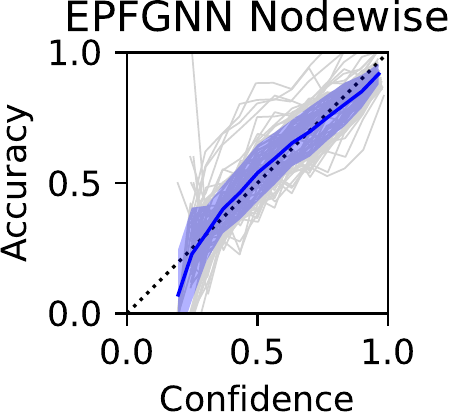} &
			\includegraphics[width=0.20\textwidth]{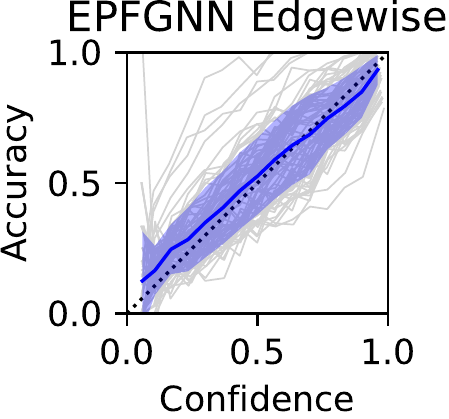} &
			\includegraphics[width=0.20\textwidth]{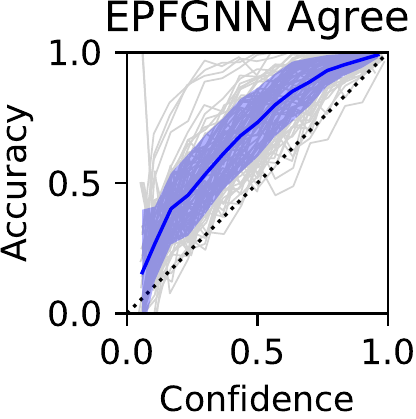} &
			\includegraphics[width=0.20\textwidth]{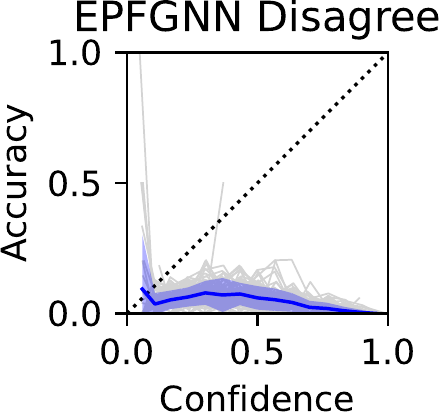} \\
		\end{tabular}
		\caption{Reliability diagrams of GCN, GAT, GMNN, and EPFGNN trained on Citeseer. The light gray lines are the results of individual experiments. The solid blue line and the shaded blue area are the mean and the standard deviation over 75 runs.} \label{fig:miscalib_tendency_citeseer}
	\end{figure}

\begin{figure}[ht]
		\centering
		\setlength{\tabcolsep}{0.02\textwidth} 
		\begin{tabular}{cccc}
			\includegraphics[width=0.20\textwidth]{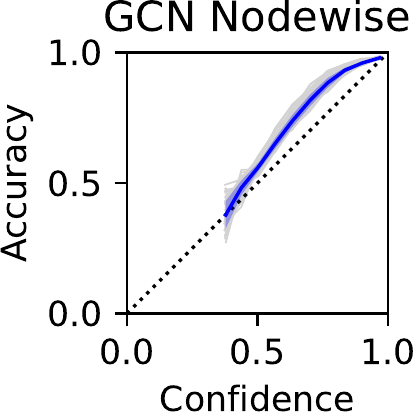} &
			\includegraphics[width=0.20\textwidth]{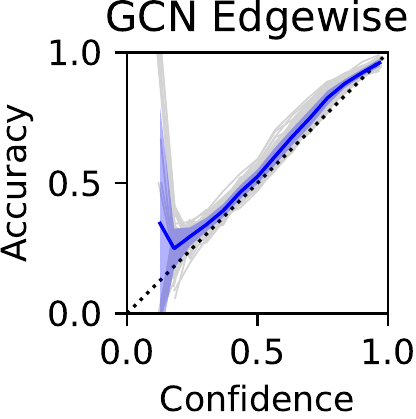} &
			\includegraphics[width=0.20\textwidth]{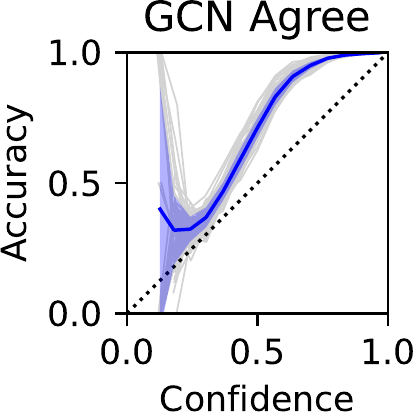} &
			\includegraphics[width=0.20\textwidth]{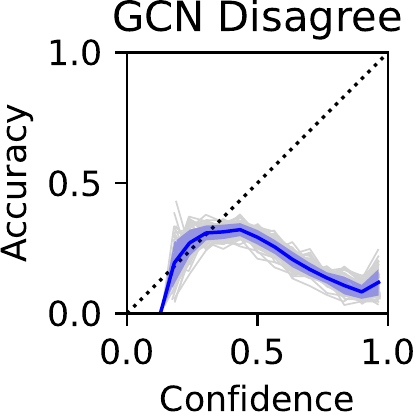} \\
			\includegraphics[width=0.20\textwidth]{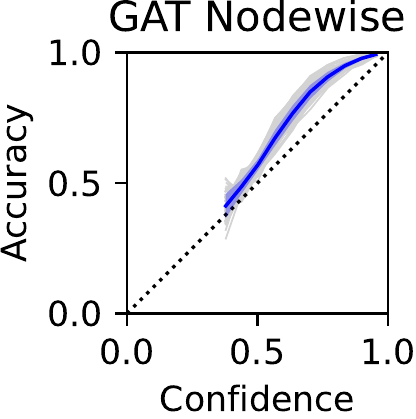} &
			\includegraphics[width=0.20\textwidth]{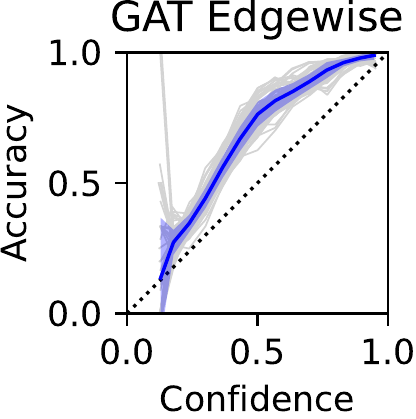} &
			\includegraphics[width=0.20\textwidth]{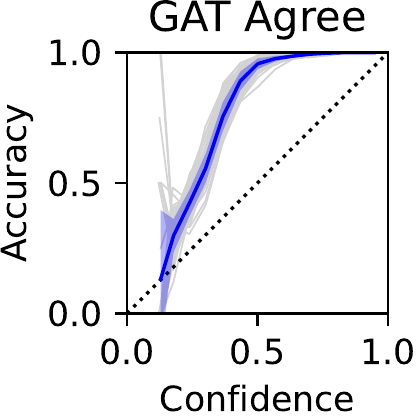} &
			\includegraphics[width=0.20\textwidth]{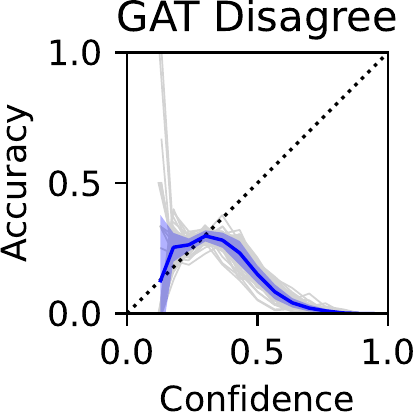} \\
			\includegraphics[width=0.20\textwidth]{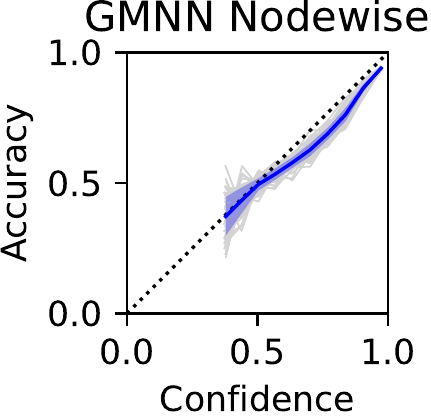} &
			\includegraphics[width=0.20\textwidth]{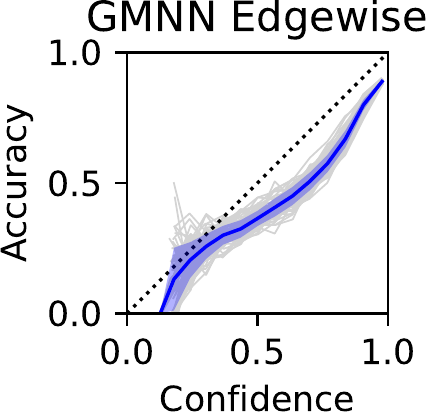} &
			\includegraphics[width=0.20\textwidth]{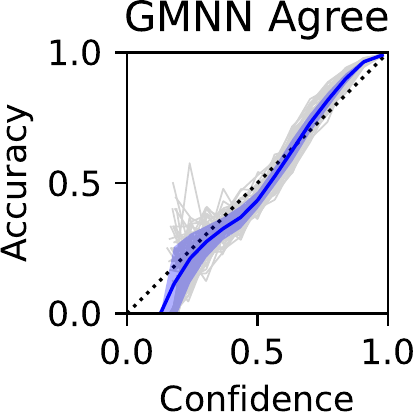} &
			\includegraphics[width=0.20\textwidth]{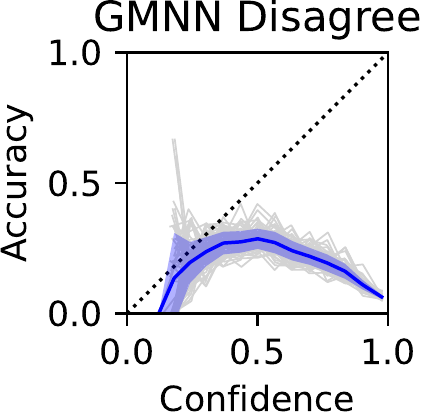} \\
			\includegraphics[width=0.20\textwidth]{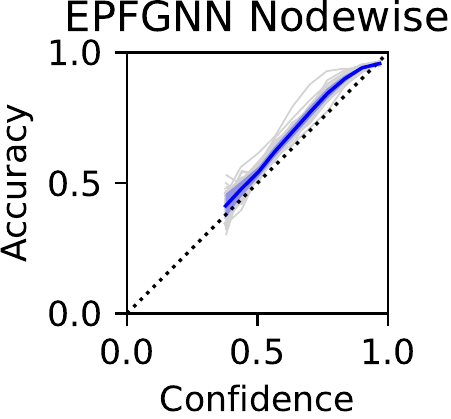} &
			\includegraphics[width=0.20\textwidth]{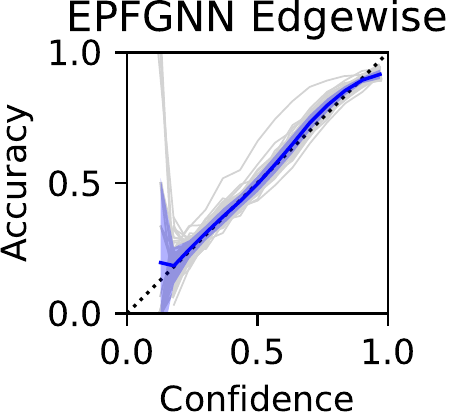} &
			\includegraphics[width=0.20\textwidth]{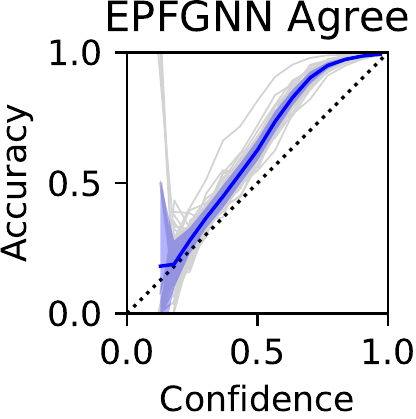} &
			\includegraphics[width=0.20\textwidth]{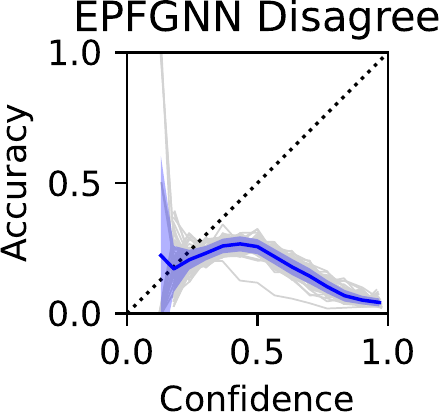} \\
		\end{tabular}
		\caption{Reliability diagrams of GCN, GAT, GMNN, and EPFGNN trained on Pubmed. The light gray lines are the results of individual experiments. The solid blue line and the shaded blue area are the mean and the standard deviation over 75 runs.} \label{fig:miscalib_tendency_pubmed}
	\end{figure}

In this section, we collect additional experimental results. Table~\ref{tab:acc_comparison} summarizes the classification results in terms of nodewise, edgewise, agree, and disagree accuracies. And Figures~\ref{fig:miscalib_tendency_cora}, \ref{fig:miscalib_tendency_citeseer} and \ref{fig:miscalib_tendency_pubmed} display the reliability diagrams for GNN predictions on Cora, Citeseer and Pubmed.

	\begin{table}[ht]
		\caption{Classification results in terms of nodewise, edgewise, agree and disagree accuracies for various GNN models and datasets. For all metrics, higher is better.}
		\label{tab:acc_comparison}
		\vspace{7pt} 
		\centering
		\small
		\begin{tabular}{lcccccc}
			\toprule
			Dataset 
			&Model 
			&Nodewise
			&Edgewise
			&Agree
			&Disagree\\
			\midrule
			\multirow{2}{*}{Cora} &GCN &82.86$\pm$0.74 &75.01$\pm$1.28 &87.28$\pm$1.36 &\textbf{23.93$\pm$2.50}\\
			& GAT &\textbf{83.71$\pm$0.73} &\textbf{75.80$\pm$1.00} &\textbf{88.40$\pm$1.20} &23.33$\pm$1.89\\
			& GMNN &82.73$\pm$0.86 &75.46$\pm$1.39 &88.33$\pm$1.35 &21.92$\pm$2.16\\
			& EPFGNN &76.48$\pm$1.16 &65.07$\pm$1.85 &76.32$\pm$1.74 &18.25$\pm$2.51\\
			\midrule
			\multirow{2}{*}{Citeseer} &GCN &\textbf{72.06$\pm$0.87} &\textbf{65.10$\pm$1.12} &87.15$\pm$1.11 &\textbf{6.59$\pm$1.49}\\
			& GAT &72.04$\pm$0.76 &64.96$\pm$1.17 &\textbf{87.59$\pm$0.87} &4.92$\pm$1.03 \\
			& GMNN &70.81$\pm$0.89 &64.49$\pm$1.16 &87.19$\pm$0.96 &4.27$\pm$1.04 \\
			& EPFGNN &69.13$\pm$1.15 &61.20$\pm$1.68 &82.45$\pm$1.95 &4.81$\pm$1.25\\
			\midrule
			\multirow{2}{*}{Pubmed} &GCN &\textbf{86.43$\pm$0.26} &\textbf{79.07$\pm$0.38} &92.98$\pm$0.24 &\textbf{22.08$\pm$0.95} \\
			& GAT &85.12$\pm$0.36 &78.38$\pm$0.47 &\textbf{93.01$\pm$0.30} &18.47$\pm$1.10\\
			& GMNN &84.23$\pm$0.29 &77.70$\pm$0.46 &92.82$\pm$0.29 &15.75$\pm$0.98\\
			& EPFGNN &82.51$\pm$0.58 &74.35$\pm$0.67 &88.40$\pm$0.65 &16.81$\pm$1.22 \\
			\bottomrule
		\end{tabular}
	\end{table}

Note the nodewise accuracies reported in Table~\ref{tab:acc_comparison} have significant discrepancies compared to the ones reported in \citet{wang2021epfgnn}. This is caused by two main factors:
\begin{itemize}
    \item At test time we employ the loopy belief propagation inference which also estimates edgewise marginals, while in \citet{wang2021epfgnn} naive mean field is used instead;
    \item We have a different experimental setup and splitting (see Appendix~\ref{asec:detail_exp_setup}).
\end{itemize}
We report the corresponding results for EPFGNN using naive mean field in Table~\ref{tab:epfgnn_mf}, which are more in line with the ones from  \citet{wang2021epfgnn}. We see that switching from naive mean field to loopy belief propagation yields better ECEs but worse accuracies. The decrease in accuracies might be explained by the mismatch of the inference method since naive mean field is used doing training.

\begin{table}[ht]
		\caption{Results for EPFGNN when naive mean field is used for test time inference instead.}
		\label{tab:epfgnn_mf}
		\vspace{7pt} 
		\centering
		\small
		\begin{tabular}{lcccccc}
			\toprule
            Dataset
            &Metric
			&Nodewise
			&Edgewise
			&Agree
			&Disagree\\
			\midrule
			\multirow{2}{*}{Cora} & Accuracy & $79.74 \pm 1.09$ & $70.02 \pm 2.00$ &$82.21\pm1.88$  &$19.30\pm2.69$  \\
			& ECE & $9.16 \pm 2.41$ & $12.53 \pm 3.31$ &$20.03\pm4.10$  &$24.07\pm4.40$  \\
			\midrule
			\multirow{2}{*}{Citeseer} & Accuracy & $71.77 \pm 0.96$ & $64.85 \pm 1.30$ &$87.56\pm1.17$  &$4.58\pm1.20$  \\
			& ECE & $8.42 \pm 6.01$ & $12.32 \pm 8.24$ &$22.53\pm13.97$  &$44.61\pm13.51$  \\
			\midrule
			\multirow{2}{*}{Pubmed} & Accuracy & $84.74 \pm 0.55$ & $76.76 \pm 0.55$ &$91.27\pm0.52$  &$17.31\pm1.20$  \\
			& ECE & $6.13 \pm 1.40$ & $3.66 \pm 1.56$ &$11.26\pm2.07$  &$44.97\pm2.23$  \\
			\bottomrule
		\end{tabular}
	\end{table}

\end{document}